\documentclass{article} % For LaTeX2e
\PassOptionsToPackage{table}{xcolor}
\usepackage{iclr2025_conference,times}

\usepackage{amsmath,amsfonts,bm}

\def\eqref#1{equation~\ref{#1}}
\def\1{\bm{1}}

\DeclareMathAlphabet{\mathsfit}{\encodingdefault}{\sfdefault}{m}{sl}
\SetMathAlphabet{\mathsfit}{bold}{\encodingdefault}{\sfdefault}{bx}{n}

\iclrfinalcopy
\usepackage{booktabs}
\usepackage{longtable}
\usepackage{array}
\usepackage{caption}

\usepackage{xcolor}
\usepackage{booktabs}
\usepackage{multirow}
\usepackage{wrapfig}
\usepackage{float}
\usepackage{placeins}
\usepackage{booktabs}
\usepackage{pifont}
\newcommand{\cmark}{\ding{51}}
\newcommand{\xmark}{\ding{55}}
\usepackage{hyperref}
\usepackage{url}
\usepackage{xspace}
\usepackage{caption}
\usepackage{bbm}
\usepackage{graphicx}

\usepackage{graphicx}
\definecolor{rankblue}{HTML}{DCEEFF}

\usepackage{booktabs}   % \toprule \midrule \cmidrule \bottomrule
\usepackage{multirow}   % \multirow
\title{DistillAlign: Coordinating Mode Covering and Mode Seeking in Autoregressive Video Distillation}
\author{
\textbf{Jiaxing Li}\textsuperscript{1,2*},
\textbf{Kai Zou}\textsuperscript{1*},
\textbf{Cindy Zhou}\textsuperscript{1,3\scalebox{0.6}{$\ddagger$}},
\textbf{Kaichen Huang}\textsuperscript{1,2},
\textbf{Junyao Gao}\textsuperscript{1}, 
 \\
\textbf{Zile Wang}\textsuperscript{1},
\textbf{Yang Liu}\textsuperscript{1},
\textbf{Bin Liu}\textsuperscript{1},
\textbf{Bo An}\textsuperscript{2},
\textbf{Yangguang Li}\textsuperscript{1$\dagger$}
\\[0.3em]
\textsuperscript{1}Riemann Dynamics,\quad
\textsuperscript{2}Nanyang Technological University \quad
\textsuperscript{3}Wellington College, UK
\\[0.25em]
}
\begin{document}

\maketitle

\renewcommand{\thefootnote}{\fnsymbol{footnote}}
\footnotetext[1]{Equal contribution.}
\footnotetext[2]{Corresponding author.}
\footnotetext[3]{Research Intern at Riemann Dynamics.}
\renewcommand{\thefootnote}{\arabic{footnote}}

\begin{abstract}
Existing autoregressive video distillation methods commonly adopt a Distribution Matching Distillation (DMD)-based multi-stage pipeline. 
% However, they typically decouple the initialization and DMD stages: the initialization matches base-model trajectories or external video data, whereas DMD refines toward a separate—often stronger—diffusion teacher, so the two stages pursue different target distributions.Intermediate students are then judged mainly by visual scores such as VBench.
However, they typically decouple the initialization and DMD stages—which then pursue different target distributions—and judge the intermediate student mainly by visual scores such as VBench.
In this paper, we revisit this design from a distributional perspective.
Given the mode-seeking nature of the distribution matching loss, a good initialization should match the mode coverage of the target DMD teacher, rather than merely pursuing high quality.
To analyze this, we introduce a distributional evaluation protocol that measures precision and coverage between student and teacher distributions in a shared latent space. It exposes differences hidden by visual scores: some initializations reach high precision but low coverage, leading to suboptimal refinement, while mode-covering ones preserve broader support.
Furthermore, even when the target distributions are aligned, DMD's reverse-KL objective can still drive the student toward high-probability teacher regions in late training, reducing coverage and diversity. To address this, we propose joint distillation, which combines DMD's mode-seeking objective with a Consistency Distillation-based mode-covering constraint.
Experiments show that our method improves generation quality, coverage, and diversity; notably, even with a Wan-1.3B DMD teacher, it outperforms baselines refined with Wan-14B, underscoring the importance of distributional alignment in autoregressive video distillation.
Project page: \url{https://lijiaxing0213.github.io/DistillAlign}
% \vspace{10pt}

\end{abstract}

\section{Introduction}
\label{intro}
Driven by the emerging vision of world models~\citep{2026matrixgame35, he2025matrix, wang2026matrix,team2026advancing} and interactive content creation~\citep{pixverse2026r1blog}, the demand for high-quality, real-time autoregressive video generation is rapidly increasing. To reduce the sampling cost of video diffusion models, recent methods commonly distill bidirectional video diffusion teachers into few-step causal generators~\citep{huang2025selfforcing,yin2025causvid,zhu2026causal,yang2025longlive}.

% Existing autoregressive video distillation methods commonly follow a multi-stage pipeline. Self Forcing~\citep{huang2025selfforcing} initializes the student with ODE distillation and then applies DMD refinement, while Causal Forcing~\citep{zhu2026causal} first trains a causal generator, compresses it into a few-step student with causal ODE or consistency distillation, and finally performs DMD. Despite different implementations, these pipelines decouple the target distributions across stages by default: pre-DMD initialization may rely on a weaker base model~\citep{huang2025selfforcing,zhu2026causal} or external video data~\citep{zhao2026causalforcingpp}, whereas DMD may use a different and stronger diffusion teacher. Existing methods~\citep{ zhao2026causalforcingpp} then select or evaluate the intermediate student mainly by visual scores such as VBench ~\citep{huang2024vbench}, implicitly assuming that a better-scoring initialization leads to a better final generator.
Existing autoregressive video distillation methods commonly follow a multi-stage pipeline. Self Forcing~\citep{huang2025selfforcing} initializes the student with ODE distillation and then applies Distribution Matching Distillation (DMD) refinement, while Causal Forcing~\citep{zhu2026causal} first trains a causal model, compresses it into a few-step student with causal ODE or Consistency Distillation (CD), and finally performs DMD training.  Despite different implementations, these pipelines often treat pre-DMD initialization and DMD refinement as separate stages with potentially different target distributions: the former may be trained on samples or trajectories from the base model~\citep{huang2025selfforcing,zhu2026causal} or external video data~\citep{zhao2026causalforcingpp}, while the latter may use a stronger diffusion teacher. Intermediate students are then judged by visual scores such as VBench~\citep{huang2024vbench}, implicitly assuming that a better-scoring initialization yields a better final generator.

\begin{wrapfigure}{r}{0.52\textwidth}
    \vspace{-0.8em}
    \centering
\includegraphics[width=0.50\textwidth]{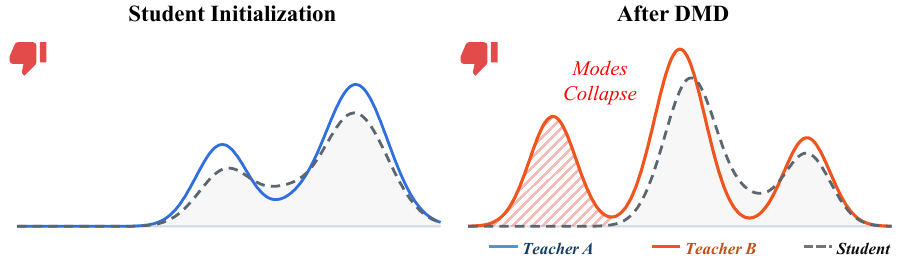}
    \vspace{-0.8em}
\captionsetup{type=figure,font=scriptsize}
    \caption{
    \textbf{Illustration of Hypothesis I.} Mismatched target distributions across stages can lead to suboptimal refinement.
    }
    \label{fig:hypothesis1}

    \vspace{-0.5em}

    \captionsetup{type=table,font=scriptsize}
    \captionof{table}{
    \textbf{Controlled target-distribution swap.}
    Built upon the Causal Forcing pipeline, we use samples from the initialization target data for causal model training and consistency distillation, and then apply DMD with different teachers.
    }
    \label{tab:toy_swap}
    % \vspace{-0.8em}

    \resizebox{0.50\textwidth}{!}{
    \begin{tabular}{cccc}
    \toprule
    Init. Target Data & DMD Teacher & Matched & VBench $\uparrow$ \\
    \midrule
    Wan-14B  & Wan-14B  & \cmark & 84.74 \\
    Wan-1.3B & Wan-1.3B & \cmark & 84.50 \\
    Wan-14B  & Wan-1.3B & \xmark & 84.16 \\
    Wan-1.3B & Wan-14B  & \xmark & 83.89 \\
    \bottomrule
    \end{tabular}
    }

    \vspace{0.2em}
    \parbox{0.50\textwidth}{\scriptsize $^{*}$The last row corresponds to the original Causal Forcing setting.}

    \vspace{-1.0em}
\end{wrapfigure}
% \begin{wrapfigure}{r}{0.50\textwidth}
% \vspace{-0.5em}
% \centering
% \includegraphics[width=0.50\textwidth]{figure/hypothesis1.pdf}
% \vspace{-2em}
% \captionsetup{font=footnotesize}
% \caption{
% Illustration of our hypothesis1.
% }
% \label{fig:hypothesis1}
% \vspace{-1.0em}
% \end{wrapfigure}

We revisit this assumption from a distribution perspective.
Since trajectory- or consistency-based initialization mainly provides mode coverage while DMD performs mode-seeking refinement around the current student distribution, DMD is effective only when its target modes are already covered by the initialization. When the target distributions of the two stages are misaligned, DMD may provide unsupported gradients, leading to coverage collapse and suboptimal refinement, as illustrated in Fig.~\ref{fig:hypothesis1}.

To verify this hypothesis, we conduct a controlled target-distribution swap on the Causal Forcing pipeline. Tab.~\ref{tab:toy_swap} shows that final performance is not determined solely by teacher capacity or data quality, but also by whether the initialization and DMD target distributions are aligned. Even weaker initialization data or a weaker DMD teacher can lead to better results when their distributions are matched. 
This suggests that initialization quality should be judged by matched mode coverage, not visual quality alone.
% This suggests that initialization quality should not be judged solely by visual quality, but also by whether it provides matched mode coverage for subsequent DMD refinement.

% \begin{wraptable}{r}{0.52\textwidth}
%     \vspace{-0.8em}
%     \centering
%     \captionsetup{font=scriptsize}
%     \caption{
%     \textbf{Controlled target-distribution swap.}
%     Built upon the Causal Forcing pipeline, we use samples from the initialization target data for causal model training and consistency distillation, and then apply DMD with different teachers.
%     }
%     \vspace{-1em}
%     \label{tab:toy_swap}
%     \resizebox{0.50\textwidth}{!}{
%     \begin{tabular}{cccc}
%     \toprule
%     Init. Target Data & DMD Teacher & Matched & VBench $\uparrow$ \\
%     \midrule
%     Wan-14B  & Wan-14B  & \cmark & 84.74 \\
%     Wan-1.3B & Wan-1.3B & \cmark & 84.50 \\
%     Wan-14B  & Wan-1.3B & \xmark & 84.16 \\
%     Wan-1.3B & Wan-14B  & \xmark & 83.89 \\
%     \bottomrule
%     \end{tabular}
%     }
%     \vspace{0.2em}
%     \parbox{0.50\textwidth}{\scriptsize $^{*}$The last row corresponds to the original Causal Forcing setting.}
%     \vspace{-1.0em}
% \end{wraptable}

Based on this finding, we propose a distributional evaluation protocol to assess the distribution matching in distillation pipelines. It projects teacher and student samples into a shared latent feature space and measures their agreement via precision and coverage. This evaluation reveals differences that are difficult to capture with VBench scores alone: some initializations achieve high visual quality and precision but low coverage, suggesting that they have collapsed to a small set of high-quality modes. In contrast, mode-covering initializations may appear blurrier, but preserve broader distributional support for subsequent DMD refinement.

\begin{wrapfigure}{r}{0.52\textwidth}
    \vspace{-0.8em}
    \centering

    \includegraphics[width=0.50\textwidth]{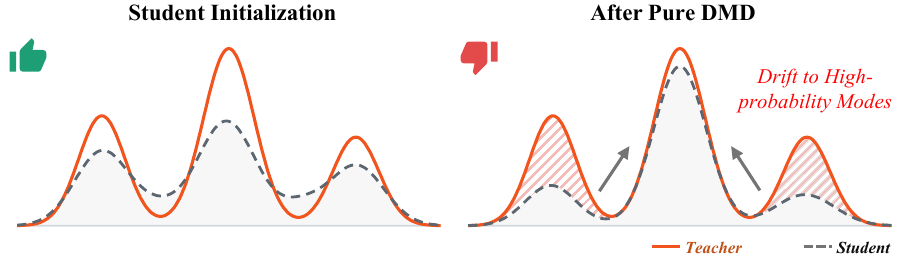}
    \vspace{-0.8em}    \captionsetup{type=figure,font=scriptsize}
    \caption{
    \textbf{Illustration of Hypothesis II.}
    Even with matched initialization, a long-term reverse-KL objective can push the student distribution toward high-probability regions of the teacher.
    }
    \label{fig:hypothesis2}

    \vspace{-0.5em}
    \captionsetup{type=table,font=scriptsize}
    \captionof{table}{
    \textbf{Late-stage drift in pure DMD training.}
    Using the first setting in Tab.~\ref{tab:toy_swap}, VBench score first increases and then decreases, while diversity consistently drops. Diversity is reported as the raw Vendi score under the protocol in Sec.~\ref{sec:setup}.
    }
    \label{tab:dmd_drift}
    % \vspace{-0.8em}

    \resizebox{0.45\textwidth}{!}{
    \begin{tabular}{lccccc}
    \toprule
    Step & 0 & 500 & 1000 & 1500 & 2500 \\
    \midrule
    VBench $\uparrow$ & 82.60 & 84.44 & 84.74 & 84.21 & 83.90 \\
    Diversity $\uparrow$ & 1.319 & 1.297 & 1.292 & 1.285 & 1.272 \\
    \bottomrule
    \end{tabular}
    }

    \vspace{-1.0em}
\end{wrapfigure}

Furthermore, even when the initialization is aligned with the DMD target, pure DMD can still suffer from late-stage distribution drift (Fig.~\ref{fig:hypothesis2}): as shown in Tab.~\ref{tab:dmd_drift}, diversity keeps decreasing while the visual score first rises and then drops, indicating that drifting toward high-probability teacher regions does not necessarily improve quality.
To address this, we propose joint distillation, which combines DMD's mode-seeking objective with CD's mode-covering constraint to sharpen high-quality modes while preserving broader teacher coverage, thereby suppressing distribution drift and proxy over-optimization.

Extensive experiments show that aligning target distributions across stages and coordinating mode-covering and mode-seeking constraints improves final generation quality and substantially enhances the student's coverage of the teacher distribution.
Notably, our method with a Wan-1.3B DMD teacher already surpasses all Wan-14B-teacher baselines, showing that distributional alignment can rival teacher scale in autoregressive video distillation.

\section{Background}
\label{background}

\subsection{Flow Matching}
% \paragraph{Flow Matching.}
Flow matching~\citep{lipman2022flow,liu2022flow} formulates generation as learning a velocity field between a simple noise distribution and the data distribution.
Given a data sample $\mathbf{x}_0 \sim p_{\mathrm{data}}$ and a noise sample $\mathbf{x}_1 \sim p_{\mathrm{noise}}$, it defines a linear interpolation path
\begin{equation}
    \label{eq:fm_interp}
    \mathbf{x}_t = (1-t)\mathbf{x}_0 + t\mathbf{x}_1,
    \quad t \in [0,1],
\end{equation}
whose target velocity is $\mathbf{x}_1-\mathbf{x}_0$.
The model is trained by minimizing
\begin{equation}
    \mathcal{L}_{\mathrm{FM}}(\theta)
    =
    \mathbb{E}_{t,\mathbf{x}_0,\mathbf{x}_1}
    \left[
    \left\|
    \mathbf{v}_{\theta}(\mathbf{x}_t,t,\mathbf{c})
    -
    (\mathbf{x}_1-\mathbf{x}_0)
    \right\|_2^2
    \right],
    \label{eq:fm_loss}
\end{equation}
where $\mathbf{c}$ denotes the conditioning signal.
During inference, samples are generated by initializing from noise and integrating the learned ODE from $t=1$ to $t=0$.

% \paragraph{Flow matching as forward KL minimization.}
% Although $\mathcal{L}_{\mathrm{FM}}$ is a pointwise regression, the
% interpolation in Eq.~\eqref{eq:fm_interp} defines a Gaussian probability path, under which
% $\mathcal{L}_{\mathrm{FM}}$ is a weighted denoising objective with a
% monotone weighting and thus admits a likelihood
% interpretation~\citep{kingma2023understanding}:
% \begin{equation}
%     \mathcal{L}_{\mathrm{FM}}(\theta)
%     =
%     -\,\mathbb{E}_{\mathbf{x}_0 \sim p_{\mathrm{data}}}
%     \big[\mathrm{ELBO}_{\theta}(\mathbf{x}_0)\big] + C
%     \;\geq\;
%     \mathrm{KL}\!\left(p_{\mathrm{data}} \,\|\, p_{\theta}\right) + C',
%     \label{eq:fm_fkl}
% \end{equation}
% where $C, C'$ are constants independent of $\theta$. 
% Although $\mathcal{L}_{\mathrm{FM}}$ is a pointwise regression, the
% interpolation in Eq.~\eqref{eq:fm_interp} defines a Gaussian probability
% path, under which $\mathcal{L}_{\mathrm{FM}}$ is a weighted denoising
% objective with a monotone weighting and thus equals the ELBO on
% noise-perturbed data~\citep{kingma2023understanding}:
% \begin{equation}
%     \mathcal{L}_{\mathrm{FM}}(\theta)
%     =
%     -\,\mathbb{E}_{t,\,\mathbf{x}_t \sim p_{\mathrm{data},t}}
%     \big[\mathrm{ELBO}_{\theta}(\mathbf{x}_t)\big] + C
%     \;\geq\;
%     \mathbb{E}_{t}
%     \big[
%     \mathrm{KL}\!\left(p_{\mathrm{data},t} \,\|\, p_{\theta,t}\right)
%     \big] + C',
%     \label{eq:fm_fkl}
% \end{equation}
% where $C, C'$ are constants independent of $\theta$. 
% Flow matching thus
% minimizes an upper bound on the forward KL divergence.

\subsection{ODE and Consistency Distillation}
ODE distillation~\citep{luhman2021knowledge,salimans2022progressive} learns a student mapping that directly predicts the teacher ODE endpoint.
Let $\Psi_{s\to r}$ denote the flow map of the teacher PF-ODE from time $s$ to $r$.
Given $\mathbf{x}_1 \sim p_{\mathrm{noise}}$ and $\mathbf{x}_t=\Psi_{1\to t}(\mathbf{x}_1)$, ODE distillation minimizes
\begin{equation}
    \mathcal{L}_{\mathrm{ODE}}(\theta)
    =
    \mathbb{E}_{t,\mathbf{x}_1}
    \left[
    \left\|
    f_{\theta}(\mathbf{x}_t,t)
    -
    \Psi_{1\to 0}(\mathbf{x}_1)
    \right\|_2^2
    \right].
    \label{eq:ode_distill}
\end{equation}

Consistency distillation (CD)~\citep{song2023consistency} instead enforces local consistency between adjacent time steps.
With a discretization $0=t_0<\cdots<t_N=1$, it uses one teacher ODE step to estimate
\begin{equation}
    \hat{\mathbf{x}}_{t_n}
    =
    \mathbf{x}_{t_{n+1}}
    +
    (t_n-t_{n+1})\,
    \mathbf{v}_{\phi}(\mathbf{x}_{t_{n+1}},t_{n+1}),
    \label{eq:cd_onestep}
\end{equation}
and trains the student by
\begin{equation}
    \mathcal{L}_{\mathrm{CD}}(\theta)
    =
    \mathbb{E}_{n,\mathbf{x}_0,\mathbf{x}_1}
    \left[
    \left\|
    f_{\theta}(\mathbf{x}_{t_{n+1}},t_{n+1})
    -
    f_{\theta^{-}}(\hat{\mathbf{x}}_{t_n},t_n)
    \right\|_2^2
    \right],
    \label{eq:cd}
\end{equation}
where $\theta^{-}$ is an EMA or stop-gradient copy of $\theta$.
At convergence, both ODE distillation and CD recover the teacher flow map,
$f_{\theta}(\mathbf{x}_t,t)=\Psi_{t\to0}(\mathbf{x}_t)$. As regressions onto teacher endpoints, such methods exhibit \emph{mode-covering} behavior, preserving broad support at the cost of sharpness.
\subsection{Distribution Matching Distillation}
Distribution matching distillation (DMD)~\citep{yin2024one,yin2024improved} distills a multi-step teacher into a few-step generator $G_{\theta}$ that maps noise directly to samples.
Let $p_{\theta,t}$ and $p_{\mathrm{teacher},t}$ denote the perturbed marginals of the generator and teacher distributions at noise level $t$.
DMD minimizes a reverse-KL objective over noise levels:
\begin{equation}
    \mathcal{L}_{\mathrm{DMD}}(\theta)
    =
    \mathbb{E}_{t}
    \left[
    w(t)\,
    \mathrm{KL}\!\left(
    p_{\theta,t}
    \,\Vert\,
    p_{\mathrm{teacher},t}
    \right)
    \right].
    \label{eq:dmd_rkl}
\end{equation}
The gradient of Eq.~\eqref{eq:dmd_rkl} can be written as a score-difference update:
\begin{equation}
    \nabla_{\theta}\mathcal{L}_{\mathrm{DMD}}
    =
    \mathbb{E}_{t,\mathbf{x}_1,\boldsymbol{\epsilon}}
    \left[
    w(t)\,
    \left(
    \mathbf{s}_{\mathrm{fake}}(\mathbf{x}_t,t)
    -
    \mathbf{s}_{\mathrm{teacher}}(\mathbf{x}_t,t)
    \right)
    \frac{\partial \mathbf{x}_t}{\partial \theta}
    \right],
    \label{eq:dmd_grad}
\end{equation}
where
$\mathbf{x}_t = (1-t)\,G_{\theta}(\mathbf{x}_1) + t\,\boldsymbol{\epsilon}$.
Here, $\mathbf{s}_{\mathrm{teacher}}$ is provided by the frozen teacher, while $\mathbf{s}_{\mathrm{fake}}$ is estimated by an online critic trained on generator samples. This reverse-KL objective is inherently \emph{mode-seeking}, concentrating the student on high-density teacher modes at the cost of coverage and diversity.
\section{Methodology}
\label{method}

\subsection{Teacher-Normalized Distribution Evaluation}
\label{sec:method_eval}

DMD sharpens the modes an initialization already covers but cannot recover those
it misses. How well the initialization covers the target teacher distribution
therefore strongly influences refinement, making this coverage---rather than
visual quality---the quantity worth measuring. Concretely, we assess the
high-level semantic distribution reached by the initializer: which scene layouts,
object configurations, motions, and teacher basins it makes accessible for later
refinement. This is non-trivial. Running a full DMD stage for every
initializer is computationally expensive, and final video quality metrics do not
directly measure distributional coverage. Directly evaluating raw initializer
outputs is also unreliable: different initializers can preserve similar
high-level semantics while differing substantially in sharpness and residual
denoising artifacts. Such low-level differences
can dominate conventional video metrics and obscure the distributional quality
of the initialization.

% \begin{figure}[H]
%     \centering
%     \includegraphics[width=0.92\linewidth]{figure/renoise_fig.pdf}
%     \caption{\textbf{Effect of teacher-normalized re-noising.}
%     We compare raw initializer samples before and after applying a shared teacher
%     refinement. Re-noising preserves the initializer-induced scene semantics
%     while reducing nuisance variation in low-level fidelity, making the
%     distributional comparison less sensitive to raw sharpness differences.}
%     \label{fig:teacher_normalized_renoise}
% \end{figure}

To address this issue, we evaluate initializers through a teacher-normalized refinement protocol. Given a prompt $c$ and noise $z$, the initializer first produces $x_{\theta}=G_{\theta}(c,z)$. We then re-noise this sample in the latent space as $x_{\theta,\rho}\sim q_{\rho}^{T_{\mathrm{norm}}}(\cdot\mid x_{\theta})$, where $q_{\rho}^{T_{\mathrm{norm}}}$ follows the forward noising process of a fixed normalization teacher $T_{\mathrm{norm}}$ to noise level $\rho$. Its solver maps the re-noised sample back to a clean video,
\begin{equation}
    \tilde{x}_{\theta}^{(\rho)}
    = R_{T_{\mathrm{norm}},\rho}(x_{\theta,\rho};c),
\end{equation}
where $R_{T_{\mathrm{norm}},\rho}$ denotes the denoising trajectory from noise level $\rho$ to a clean sample. Since all initializers are evaluated with the same re-noising level, normalization teacher, and denoising schedule, low-level fidelity differences are largely normalized, while the result still reflects the semantic region reached by the initializer, as shown in Fig.~\ref{fig:teacher_normalized_renoise}. The normalization teacher and the teacher that defines the reference distribution play distinct roles; full implementation details are provided in Appendix~\ref{app:coverage_metrics}.

We then compare refined initializer samples with teacher reference samples in a
frozen video representation space. Let $\phi(\cdot)$ be a video encoder, and
define normalized features
\begin{equation}
    u_i =
    \frac{\phi(\tilde{x}_{\theta,i}^{(\rho)})}
         {\|\phi(\tilde{x}_{\theta,i}^{(\rho)})\|_2},
    \qquad
    v_j =
    \frac{\phi(x_{T,j})}
         {\|\phi(x_{T,j})\|_2}.
\end{equation}
We estimate teacher support using a $k$-nearest-neighbor radius. Let $r_k(v_j)$
be the distance from teacher feature $v_j$ to its $k$-th nearest neighbor among
teacher features. Precision measures whether refined samples lie inside this
support:
\begin{equation}
    \mathrm{Precision}
    =
    \frac{1}{N}
    \sum_{i=1}^{N}
    \mathbf{1}
    \left[
        \exists j \;\; \mathrm{s.t.} \;\;
        \|u_i - v_j\|_2 \le r_k(v_j)
    \right].
\end{equation}
Coverage measures how much teacher support is reached by the initializer:
\begin{equation}
    \mathrm{Coverage}
    =
    \frac{1}{M}
    \sum_{j=1}^{M}
    \mathbf{1}
    \left[
        \min_i \|u_i - v_j\|_2 \le r_k(v_j)
    \right].
\end{equation}

\subsection{Mode Alignment Matters}
\subsubsection{Rethinking the Toy Experiment}
In Sec.~\ref{intro}, the controlled teacher-source swap shows that matched initialization and DMD targets lead to better final VBench scores.
Using our distributional protocol, we revisit these settings and quantify the effect under the same conditions.
Specifically, for each row, we compute coverage after initialization and after DMD, using the corresponding DMD teacher as the target distribution.
As shown in Tab.~\ref{tab:teacher_source_swap}, matched settings achieve higher coverage than mismatched ones both before and after DMD.
This confirms that the advantage of aligned stages does not merely come from stronger teachers or higher-quality data; instead, matched initialization provides better mode coverage over the DMD target distribution, making subsequent DMD refinement better supported.
In addition, we observe that DMD generally reduces coverage, which is consistent with its mode-seeking behavior. More details are provided in Sec.~\ref{sec:joint_ablation}.

\begin{table}[!htbp]
    \centering
    \captionsetup{font=footnotesize}
    \caption{\textbf{Controlled teacher-source swap.} For each setting in Table~\ref{tab:toy_swap}, we recompute the coverage of both the initialization and DMD-refined results against the corresponding teacher distribution. In each of the three numeric columns, darker shading indicates a higher-ranked value.}
    % Built upon the Causal Forcing pipeline, we use samples from the
    % initialization source for causal model training and consistency distillation,
    % and then apply DMD with different teachers. Coverage is computed against the
    % target DMD teacher in each row.}
    \label{tab:teacher_source_swap}
    \footnotesize
    \setlength{\tabcolsep}{3.8pt}
    \renewcommand{\arraystretch}{0.95}
    \begin{tabular}{cccccc}
    \toprule
    \multirow{2}{*}{Init. Data Source}
    & \multirow{2}{*}{DMD Teacher}
    & \multirow{2}{*}{Matched}
    & \multirow{2}{*}{VBench $\uparrow$}
    & \multicolumn{2}{c}{Coverage $\uparrow$} \\
    \cmidrule(lr){5-6}
    & & & & Init. & After DMD \\
    \midrule
    Wan-14B  & Wan-14B  & \cmark & \cellcolor{rankblue!100} 84.74 & \cellcolor{rankblue!100} 0.648 & \cellcolor{rankblue!100} 0.527 \\
    Wan-1.3B & Wan-1.3B & \cmark & \cellcolor{rankblue!75} 84.50 & \cellcolor{rankblue!75} 0.516 & \cellcolor{rankblue!75} 0.461 \\
    Wan-14B  & Wan-1.3B & \xmark & \cellcolor{rankblue!50} 84.16 & \cellcolor{rankblue!50} 0.503 & \cellcolor{rankblue!25} 0.430 \\
    Wan-1.3B & Wan-14B  & \xmark & \cellcolor{rankblue!25} 83.89 & \cellcolor{rankblue!25} 0.453 & \cellcolor{rankblue!50} 0.435 \\
    \bottomrule
    \end{tabular}

    \vspace{0.15em}
    {\scriptsize $^{*}$The last row corresponds to the original Causal Forcing setting.}
\end{table}
\vspace{-1.2mm}

\subsubsection{Rethinking Distillation Pipes from a Distributional Perspective}

\begin{figure}[H]
\centering
\includegraphics[width=0.95\linewidth]{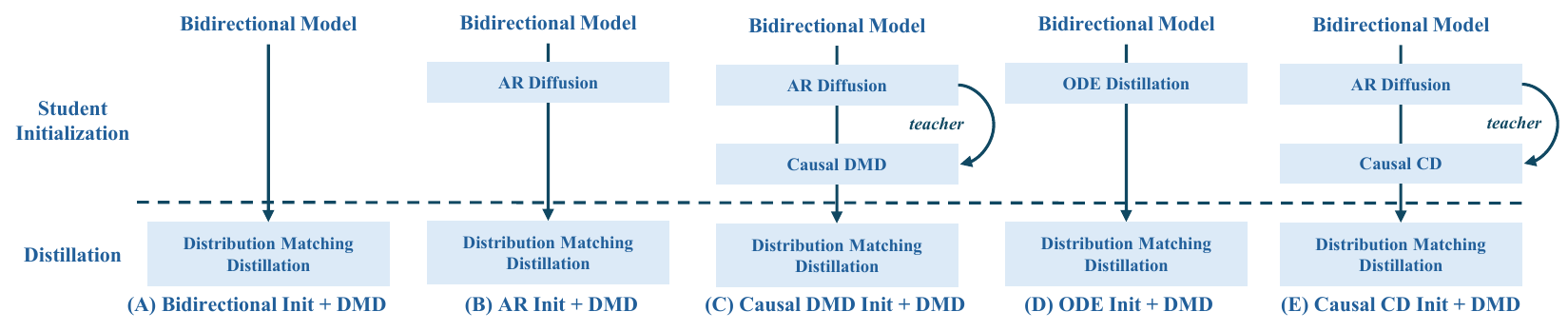}

\vspace{0.4em}

\includegraphics[width=0.95\linewidth]{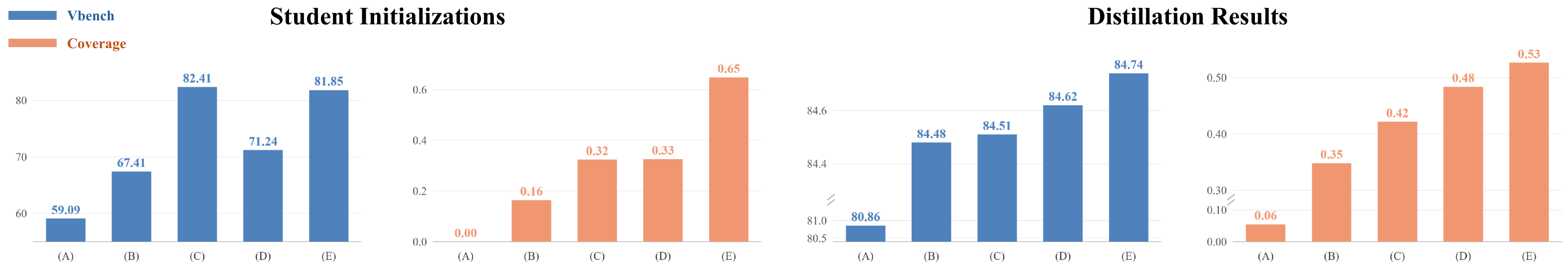}
\captionsetup{font=footnotesize}
\vspace{-0.4em}
\caption{
\textbf{Comparison across different distillation pipelines.}
All pipelines use Wan-14B as the DMD teacher and use its sampled distillation data as the initialization target.
}
\label{fig:pipeline_compare}
\end{figure}
\vspace{-1em}

When the data source and target distribution are already aligned, how do different initialization methods affect coverage and subsequent DMD refinement? 

As shown in Fig.~\ref{fig:pipeline_compare}, we construct five distillation pipelines for comparison. Among them, ODE Init + DMD and Causal CD Init + DMD correspond to the pipelines used in Self Forcing and Causal Forcing, respectively. In addition, we introduce three controlled settings: initialization directly from the bidirectional base model, initialization with AR diffusion, and initialization with AR diffusion followed by Causal DMD. The last setting forms a direct comparison with Causal CD Init: both use the Stage-1 causal model as the teacher, but Causal DMD is more mode-seeking, whereas consistency distillation is more mode-covering.

\begin{figure}[!t]
    \centering
    \includegraphics[width=0.95\linewidth]{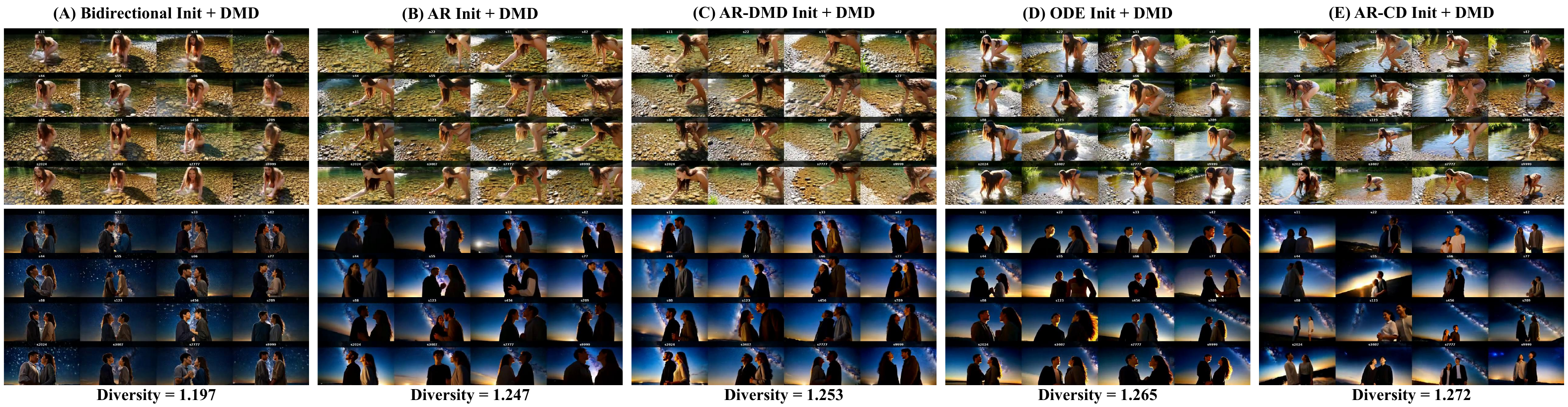}
    \caption{
    \textbf{Visual diversity comparison across different distillation pipelines.}
For each pipeline, we sample 16 fixed seeds and visualize the middle frame of each generated 81-frame video as a grid.
}
    \label{fig:pipeline_diversity}
\end{figure}

We compare the initialization and DMD-refined results of different pipelines. Except for Bidirectional Init + DMD, most AR-based pipelines achieve high VBench scores after DMD, making visual scores alone insufficient to distinguish initialization quality. Causal DMD Init is a representative case: it obtains the highest initialization VBench but does not yield the best final result. Our distribution evaluation explains this gap: AR Init and Causal DMD Init achieve high precision but low coverage, indicating concentration on limited high-quality modes. In contrast, Causal CD Init maintains broader coverage before and after DMD, providing better support for mode-seeking refinement. 
Moreover, under the same initialization target, coverage differences are also reflected in sample diversity. As shown in Fig.~\ref{fig:pipeline_diversity}, pipelines D and E exhibit higher diversity, consistent with their broader coverage.
% In addition, under the same data source, sample diversity provides an intuitive reflection of distributional coverage.

% \begin{center}
%     \includegraphics[width=1.0\linewidth]{figure/beyond_seeking_1.pdf}
%     \captionof{figure}{
%     \textbf{Distribution evolution under teacher-normalized evaluation.}
% }
%     \label{fig:beyond_seeking_pipe}
% \end{center}

% \begin{center}
%     \includegraphics[width=1.0\linewidth]{figure/beyond_seeking_2.pdf}
%     \captionof{figure}{
%     \textbf{Distribution evolution under teacher-normalized evaluation.}
% }
%     \label{fig:beyond_seeking_vis}
% \end{center}

% \clearpage

\subsection{Balancing Mode Coverage and Mode Seeking}
\label{mcms}
\subsubsection{Distribution Evolution}
\begin{figure}[!htbp]
    \centering
    \includegraphics[width=1.0\linewidth]{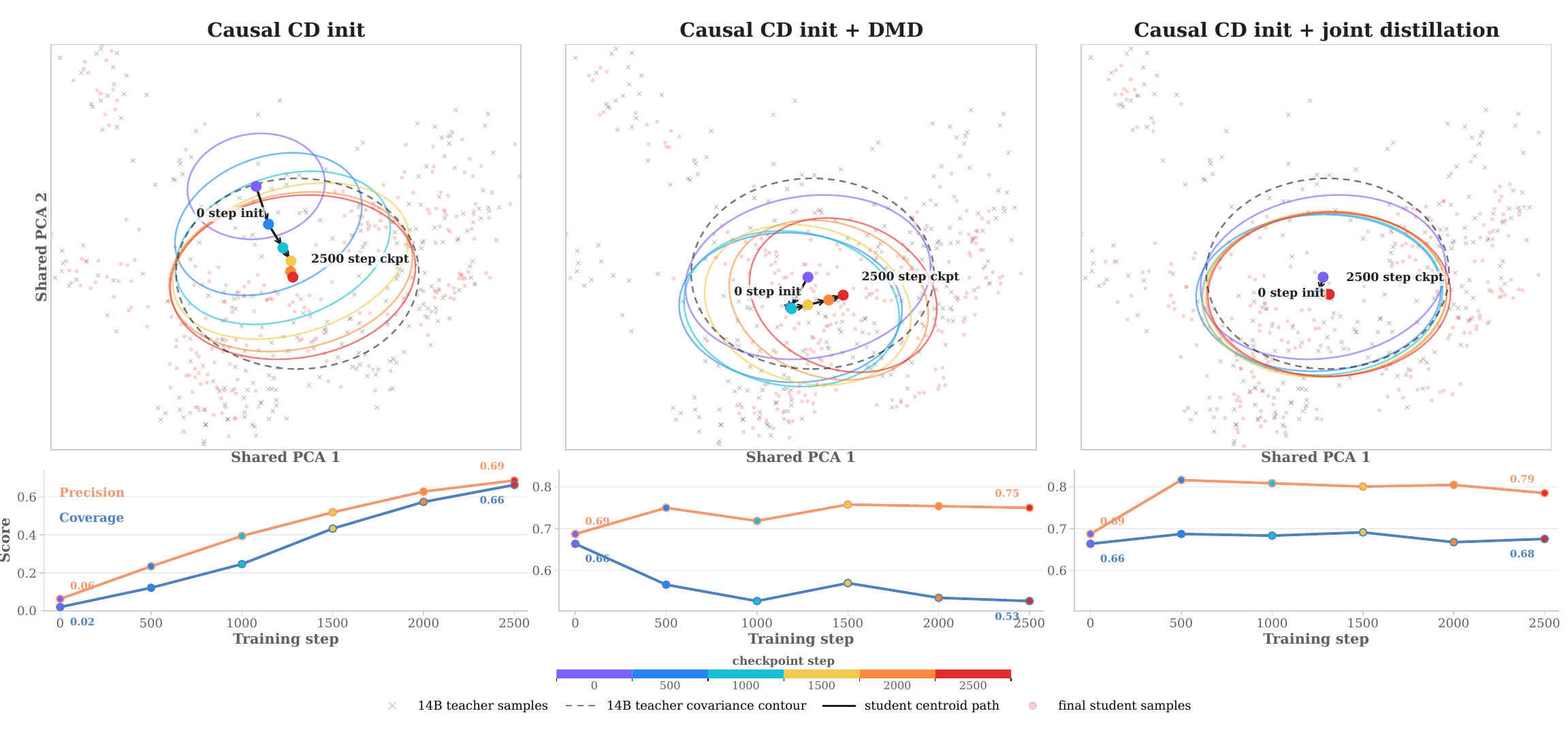}
    \caption{\textbf{Distribution evolution under mode-covering and
    mode-seeking objectives.}
    We visualize checkpoint distributions in a shared V-JEPA2 PCA space and
    track precision and coverage over training. Causal CD expands teacher
    coverage, pure DMD contracts toward high-density teacher regions, and joint
    distillation preserves coverage while refining samples.}
    \label{fig:coverage_evolution}
\end{figure}

Whereas the previous subsection compares student results \emph{across} stages, here we track the dynamics \emph{within} a single matched run, where the initialization and distillation targets share the same distribution. To
visualize how the student distribution evolves during training, we project
V-JEPA2 features onto a shared PCA basis fitted from teacher samples. In
Fig.~\ref{fig:coverage_evolution}, black crosses denote individual teacher
samples, the dashed black covariance contour summarizes the teacher distribution,
colored covariance contours show the student distribution at different
checkpoints, and light red points indicate the final student samples.

The left panel shows the evolution of the Causal CD initializer. At early
checkpoints, the student distribution is still far from the teacher distribution,
as indicated by the early blue contour being separated from the dashed teacher
contour. As training proceeds, the few-step generator gradually expands its
support and moves closer to the teacher distribution. This trend indicates the
mode-covering behavior of trajectory- or consistency-based initialization: the initializer progressively reaches more teacher modes and improves coverage more than perceptual fidelity: the samples can remain blurry or under-refined, consistent with the averaging effect of mode-covering objectives. The middle panel shows the subsequent DMD stage initialized from the matched
Causal CD model. Unlike the initialization stage, DMD no longer primarily expands
coverage. Instead, the student distribution gradually drifts toward
high-density regions of the teacher distribution, exhibiting a mode-seeking
behavior. This is consistent with the RKL-style refinement illustrated in
Fig.~\ref{fig:hypothesis2}: DMD tends to select and sharpen
high-probability modes around the current student distribution. As a result,
later DMD checkpoints may improve local fidelity and sharpness while reducing
coverage and diversity.

This drift has two implications. On the one hand, concentrating probability mass
near high-density teacher regions can improve visual quality by sharpening
samples that are already close to plausible teacher modes. On the other hand,
high-density regions under the teacher distribution do not necessarily contain
all desirable videos. Samples with large motion, rich details, or rare dynamics
may lie in relatively lower-density regions. Overly strong mode-seeking refinement can therefore suppress these valuable modes, trading diversity---and sometimes overall quality---for local realism.

\subsubsection{Joint Distillation} To mitigate the late-stage distribution drift of DMD, we combine the mode-seeking objective with a mode-covering constraint under the same target distribution:
\[
\mathcal{L}_{\mathrm{joint}}
=
\mathcal{L}_{\mathrm{DMD}}
+
\lambda \mathcal{L}_{\mathrm{CD}} .
\]
Here, $\mathcal{L}_{\mathrm{CD}}$ is the second-stage consistency distillation loss, which acts as a distributional anchor for the student. While DMD selects and sharpens high-probability modes, the CD term helps maintain coverage over the teacher distribution and prevents excessive mode collapse. As shown in Fig.~\ref{fig:coverage_evolution}(c), the joint objective enables the student distribution to match teacher high-probability regions while preserving broader coverage, thus balancing mode seeking against mode covering and suppressing the drift of pure DMD.

\section{Experiments}
\label{sec:exp}

\subsection{Setup}
\label{sec:setup}
\paragraph{Implementation details.} We adopt Wan2.1-T2V-1.3B~\citep{wang2023videolcm} as the base model for finetuning, which generates 81-frame videos at a resolution of $832\times480$. In the main experiments, we build upon the three-stage pipeline of Causal Forcing. We collect two sets of distillation data from Wan2.1-1.3B and Wan2.1-14B, respectively, each containing 25K samples. The prompt list is from VidProM~\citep{wang2024vidprom}. In the first stage, we perform autoregressive diffusion training with teacher forcing for 5K steps on each data set. In the second stage, we apply consistency distillation for 2.5K steps to obtain a few-step causal generator. In the final stage, we conduct 1.5K steps of joint DMD-CD training. The DMD teacher is chosen to match the corresponding data source, while the CD teacher is kept the same as the causal teacher used in Stage 2. We set the joint loss weight $\lambda$ to $0.01$.

% \paragraph{Datasets.}

\paragraph{Evaluation.}
We evaluate perceptual quality with the VBench text-to-video protocol
\citep{huang2024vbench}, reporting quality, semantic, and weighted overall scores
under the same prompt set and scripts for all methods. To measure distributional behavior, we use the
teacher-normalized protocol introduced in Sec.~\ref{sec:method_eval}. Unless
otherwise specified, we use $N=M=256$ student and teacher samples on matched
prompts and seeds. The teacher reference set is generated by the target teacher
listed in each row. By default, we extract V-JEPA2 features from 8 uniformly
sampled frames within the first 5 seconds of each video and compute precision
and coverage with a $k$-NN support estimator using $k=5$. The dedicated
first-chunk analysis in Fig.~\ref{fig:pipeline_compare} and
Tab.~\ref{tab:pipeline_full} instead uses the first 12 consecutive frames, as
detailed in Appendix~\ref{app:coverage_metrics}. For initialization-stage models, we first apply teacher-normalized refinement ($\rho=0.9$) and then compute precision and coverage against the target teacher; for post-DMD models, we evaluate the final samples directly. When DMD teachers differ, precision and coverage are computed against the corresponding teacher and should be read relative to its support. For
non-DMD systems that do not define a target DMD teacher, we omit
teacher-normalized precision and coverage when appropriate. Diversity is reported
as the raw Vendi score~\citep{friedman2023vendi} computed from the same
normalized V-JEPA2 feature similarities.

\FloatBarrier

\subsection{Main Comparison}
\label{sec:comparison}
Table~\ref{tab:main_comparison} compares our method with both open few-step video generation systems and autoregressive diffusion distillation baselines. The open baselines include LTX Video, Wan2.1-T2V-1.3B, SkyReels-DF, and CausVid, evaluated with the same VBench protocol. For DMD-based autoregressive methods, we additionally report the DMD teacher, which separates the effect of teacher capacity from the effect of teacher-aligned initialization.

% \clearpage

\begin{table}[!htbp]
\centering
\caption{
Main comparison on text-to-video generation. VBench is reported with total,
quality, and semantic scores. Precision and coverage are
teacher-normalized against the listed DMD teacher when applicable; diversity is
the raw Vendi score. ``DMD Teacher'' is only applicable to methods
with a DMD refinement stage.
}
\label{tab:main_comparison}
\small
\setlength{\tabcolsep}{2.3pt}
\begin{tabular}{lccccccc}
\toprule
\multirow{2}{*}{Method} & \multirow{2}{*}{DMD Teacher}
& \multicolumn{3}{c}{VBench Metrics $\uparrow$}
& \multicolumn{3}{c}{Distributional Metrics $\uparrow$} \\
\cmidrule(lr){3-5}\cmidrule(lr){6-8}
& & Total & Quality & Semantic & Precision & Coverage & Diversity \\
\midrule
\multicolumn{8}{l}{\textit{Bidirectional models}} \\
LTX Video & -- & 81.37 & 83.05 & 74.66 & -- & -- & 1.281 \\
Wan2.1-T2V-1.3B    & -- & 83.56 & 84.31 & 80.55 & -- & -- & 1.334 \\
% \midrule
% \multicolumn{8}{l}{\textit{Autoregressive models}} \\
% %NOVA        & -- & 78.96 & 74.93 & 78.15 & 0.145 & 0.074 & -- \\
% SkyReels-DF & -- & 82.88 & 76.25 & 81.56 & -- & -- & 1.335 \\
\midrule
\multicolumn{8}{l}{\textit{Distilled autoregressive models}} \\
CausVid   & Wan2.1-14B & 83.03 & 83.99 & 79.18 & 0.51 & 0.21 & 1.223 \\
%LongLive  & -- & 82.30 & 82.77 & 80.39 & 0.598 & 0.301 & -- \\
Self-Forcing   & Wan2.1-14B  & 84.21 & 85.04 & \textbf{80.87} & \underline{0.78} & 0.53 & 1.269 \\
Causal-Forcing & Wan2.1-14B  & 84.38 & 85.47 & 80.01 & 0.61 & 0.29 & 1.250 \\
Ours (weak)    & Wan2.1-1.3B & \underline{84.54} & \underline{85.61} & 80.28 & 0.77 & \underline{0.59} & \textbf{1.305} \\
Ours (full)    & Wan2.1-14B  & \textbf{84.83} & \textbf{85.92} & \underline{80.47} & \textbf{0.79} & \textbf{0.69} & \underline{1.304} \\
\bottomrule
\end{tabular}
\end{table}

Our method achieves the best overall VBench score among the compared methods and
also obtains the strongest coverage among distilled autoregressive models.
Importantly, the comparison separates two factors that are often conflated in
DMD-based autoregressive distillation: the DMD teacher and the initialization
distribution. The weak variant remains competitive even when refined with a
smaller Wan2.1-1.3B teacher, while the full variant further improves performance
with the Wan2.1-14B teacher. This supports the view developed in
Sec.~\ref{method}: DMD benefits from an initialization that already covers the
target teacher distribution, rather than relying on the refinement stage to
recover missing modes. The coverage improvement is not merely a diagnostic
artifact: it is accompanied by stronger raw diversity and better final VBench
under the matched DMD teacher.

\begin{figure}[!htbp]
  \centering
  \includegraphics[width=\linewidth]{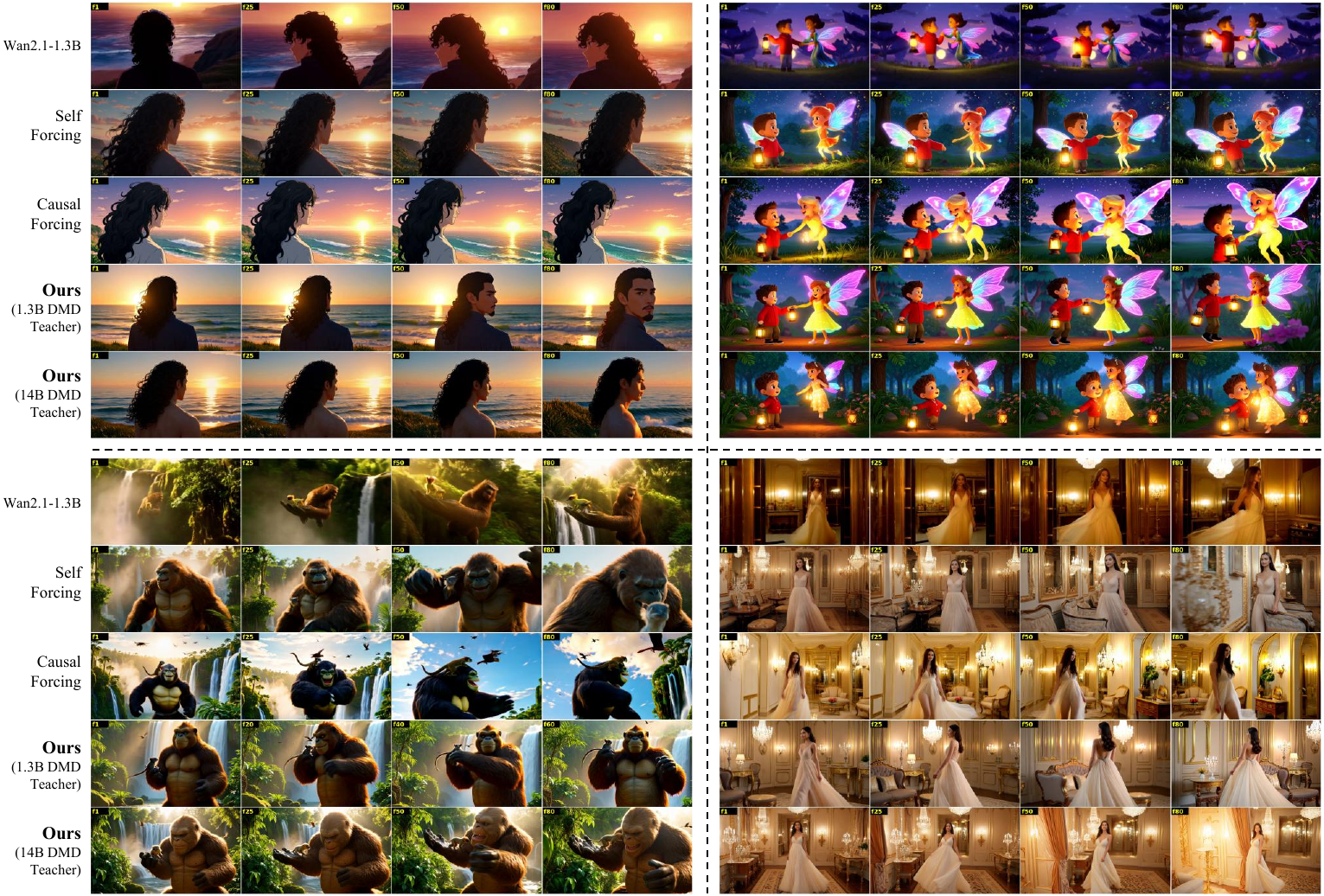}
  \vspace{-6mm}
  \caption{\textbf{Qualitative comparison of visual quality and motion.}
  Compared with representative bidirectional and distilled autoregressive
  baselines, our method produces sharper subjects, stronger motion, and more
  temporally coherent details.}
  \label{fig:exp_comparison}
   % \vspace{-6mm}
\end{figure}

Fig.~\ref{fig:exp_comparison} provides qualitative examples consistent with the
quantitative trend. Compared with prior distilled autoregressive baselines, our
model better preserves object identity and motion details across frames, while
avoiding the overly smoothed appearance typical of mode-covering initialization
alone.

\begin{figure}[!htbp]
  \centering
  \includegraphics[width=\linewidth]{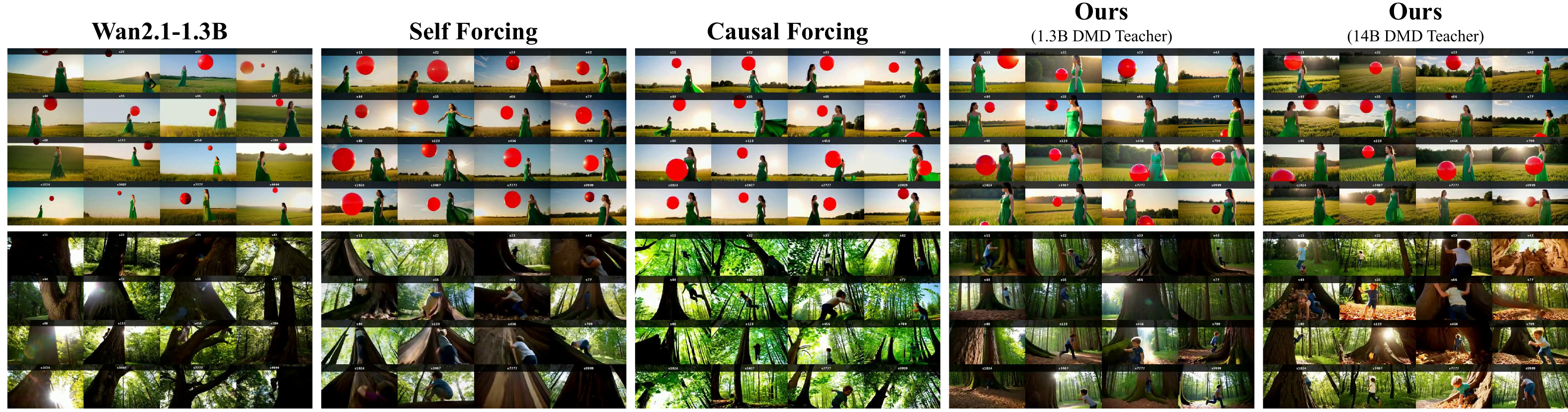}
  \vspace{-6mm}
  \caption{\textbf{Qualitative comparison of sample diversity.}
  Under fixed prompts and different seeds, our method preserves richer
  appearance and motion variations while maintaining high visual fidelity.}
  \label{fig:exp_comparison_diversity}
   \vspace{-3mm}
\end{figure}

Fig.~\ref{fig:exp_comparison_diversity} further visualizes the diversity
difference across seeds. Methods with lower teacher coverage tend to produce
more similar appearances or motion patterns, whereas our matched joint
distillation preserves broader variations without sacrificing visual quality.

% \FloatBarrier

% \subsection{Supplementary Ablation}
% \label{sec:ablation}
% \subsubsection{Complete Results of}

\subsection{Ablations and Additional Results}
\subsubsection{Joint Distillation}
\label{sec:joint_ablation}
As a complement to the analysis in Sec.~\ref{mcms}, we ablate joint distillation from two aspects: the joint loss weight $\lambda$, and the training dynamics across steps.

We first ablate the joint loss weight $\lambda$, which balances the DMD mode-seeking objective against the CD-based mode-covering constraint. For each $\lambda$, we train for 2500 steps in total and compare the best checkpoint along the training trajectory. As shown in Tab.~\ref{tab:lambda_ablation}, a larger $\lambda$ emphasizes the CD-based mode-covering constraint, leading to the highest coverage but lower VBench scores. As $\lambda$ decreases, the DMD mode-seeking effect becomes stronger, improving visual quality and precision, and $\lambda=0$ reduces to pure DMD. Overall, the best trade-off among VBench, precision, and coverage is achieved at an intermediate $\lambda$; in our experiments this optimum is $\lambda=0.01$, which we adopt as our default setting.

\begin{table}[t]
\centering
\captionsetup{font=footnotesize}
\caption{
\textbf{Ablation on the joint loss weight $\lambda$.}
A larger $\lambda$ strengthens the CD-based mode-covering constraint, while a smaller $\lambda$ gives more weight to DMD refinement, and $\lambda=0$ recovers pure DMD. For each $\lambda$, we report the best checkpoint within a 2500-step training trajectory. Diversity is reported as the raw Vendi score under the same V-JEPA2 protocol as Tab.~\ref{tab:main_comparison}. In each column, darker shading indicates a larger value.
}
\label{tab:lambda_ablation}
\footnotesize
\setlength{\tabcolsep}{5.0pt}
\renewcommand{\arraystretch}{1.05}
\begin{tabular}{c ccc ccc}
\toprule
\multirow{2}{*}{$\lambda$}
& \multicolumn{3}{c}{VBench Metrics $\uparrow$}
& \multicolumn{3}{c}{Distributional Metrics $\uparrow$} \\
\cmidrule(lr){2-4}\cmidrule(lr){5-7}
& Total & Quality & Semantic & Precision & Coverage & Diversity \\
\midrule
1
& \cellcolor{rankblue!14} 84.18 & \cellcolor{rankblue!14} 85.17 & \cellcolor{rankblue!14} 80.18 & \cellcolor{rankblue!32} 0.73 & \cellcolor{rankblue!100} 0.71 & \cellcolor{rankblue!100} 1.311 \\
0.1
& \cellcolor{rankblue!32} 84.67 & \cellcolor{rankblue!32} 85.72 & \cellcolor{rankblue!32} 80.46 & \cellcolor{rankblue!56} 0.76 & \cellcolor{rankblue!80} 0.69 & \cellcolor{rankblue!32} 1.303 \\
0.01
& \cellcolor{rankblue!100} 84.83 & \cellcolor{rankblue!100} 85.92 & \cellcolor{rankblue!56} 80.47 & \cellcolor{rankblue!100} 0.79 & \cellcolor{rankblue!80} 0.69 & \cellcolor{rankblue!56} 1.304 \\
0.001
& \cellcolor{rankblue!80} 84.78 & \cellcolor{rankblue!80} 85.85 & \cellcolor{rankblue!100} 80.55 & \cellcolor{rankblue!80} 0.78 & \cellcolor{rankblue!56} 0.64 & \cellcolor{rankblue!80} 1.307 \\
0
& \cellcolor{rankblue!56} 84.74 & \cellcolor{rankblue!56} 85.78 & \cellcolor{rankblue!80} 80.54 & \cellcolor{rankblue!14} 0.72 & \cellcolor{rankblue!32} 0.53 & \cellcolor{rankblue!14} 1.292 \\
\bottomrule
\end{tabular}

\vspace{0.2em}
\end{table}

We then examine the behavior of joint distillation across training steps under the default $\lambda=0.01$, quantifying the distribution-evolution trend visualized in Fig.~\ref{fig:coverage_evolution}.
As shown in Tab.~\ref{tab:joint_ablation}, pure DMD improves visual quality in the early stage, but its coverage and diversity continuously decrease as training proceeds, indicating that mode-seeking refinement gradually contracts the student distribution.
In contrast, joint distillation maintains higher coverage and diversity throughout training, while achieving comparable or even better quality, semantic, and total scores.
In the later training stage, joint distillation preserves substantially higher coverage and diversity than pure DMD and also obtains a higher total score.
This shows that introducing the CD-based mode-covering constraint effectively suppresses DMD-induced distribution drift without sacrificing generation quality.

\begin{table*}[t]
\centering
\captionsetup{font=footnotesize}
\caption{
Ablation of joint distillation across training steps.
Compared with pure DMD, joint distillation better preserves coverage and diversity while achieving comparable or better VBench scores. Diversity is reported as the raw Vendi score under the same V-JEPA2 protocol as Tab.~\ref{tab:main_comparison}. In each column, darker shading indicates a larger value.
}
\label{tab:joint_ablation}
\scriptsize
\setlength{\tabcolsep}{3.0pt}
\renewcommand{\arraystretch}{1.08}
\resizebox{\textwidth}{!}{
\begin{tabular}{c ccc ccc @{\hspace{0.8em}} ccc ccc}
\toprule
\multirow{3}{*}{Step}
& \multicolumn{6}{c}{Joint Distillation ($\lambda=0.01$)}
& \multicolumn{6}{c}{Pure DMD ($\lambda=0$)} \\
\cmidrule(lr){2-7}\cmidrule(lr){8-13}
& \multicolumn{3}{c}{VBench Metrics $\uparrow$} & \multicolumn{3}{c}{Distributional Metrics $\uparrow$}
& \multicolumn{3}{c}{VBench Metrics $\uparrow$} & \multicolumn{3}{c}{Distributional Metrics $\uparrow$} \\
\cmidrule(lr){2-4}\cmidrule(lr){5-7}\cmidrule(lr){8-10}\cmidrule(lr){11-13}
& Total & Quality & Semantic & Precision & Coverage & Diversity
& Total & Quality & Semantic & Precision & Coverage & Diversity \\
\midrule
0
& \cellcolor{rankblue!14} 82.60 & \cellcolor{rankblue!14} 84.18 & \cellcolor{rankblue!14} 76.27 & \cellcolor{rankblue!32} 0.69 & \cellcolor{rankblue!32} 0.66 & \cellcolor{rankblue!100} 1.319
& \cellcolor{rankblue!14} 82.60 & \cellcolor{rankblue!14} 84.18 & \cellcolor{rankblue!14} 76.27 & \cellcolor{rankblue!32} 0.69 & \cellcolor{rankblue!100} 0.66 & \cellcolor{rankblue!100} 1.319 \\

500
& \cellcolor{rankblue!56} 84.60 & \cellcolor{rankblue!80} 85.87 & \cellcolor{rankblue!32} 79.53 & \cellcolor{rankblue!100} 0.82 & \cellcolor{rankblue!56} 0.67 & \cellcolor{rankblue!56} 1.305
& \cellcolor{rankblue!80} 84.44 & \cellcolor{rankblue!80} 85.52 & \cellcolor{rankblue!80} 80.09 & \cellcolor{rankblue!80} 0.75 & \cellcolor{rankblue!80} 0.56 & \cellcolor{rankblue!80} 1.297 \\

1000
& \cellcolor{rankblue!80} 84.78 & \cellcolor{rankblue!56} 85.85 & \cellcolor{rankblue!100} 80.52 & \cellcolor{rankblue!80} 0.80 & \cellcolor{rankblue!80} 0.68 & \cellcolor{rankblue!32} 1.304
& \cellcolor{rankblue!100} 84.74 & \cellcolor{rankblue!100} 85.78 & \cellcolor{rankblue!100} 80.54 & \cellcolor{rankblue!56} 0.72 & \cellcolor{rankblue!32} 0.53 & \cellcolor{rankblue!56} 1.292 \\

1500
& \cellcolor{rankblue!100} 84.83 & \cellcolor{rankblue!100} 85.92 & \cellcolor{rankblue!80} 80.47 & \cellcolor{rankblue!56} 0.79 & \cellcolor{rankblue!100} 0.69 & \cellcolor{rankblue!32} 1.304
& \cellcolor{rankblue!56} 84.40 & \cellcolor{rankblue!56} 85.50 & \cellcolor{rankblue!56} 80.01 & \cellcolor{rankblue!100} 0.76 & \cellcolor{rankblue!56} 0.55 & \cellcolor{rankblue!32} 1.285 \\

2500
& \cellcolor{rankblue!32} 84.27 & \cellcolor{rankblue!32} 85.26 & \cellcolor{rankblue!56} 80.32 & \cellcolor{rankblue!56} 0.79 & \cellcolor{rankblue!80} 0.68 & \cellcolor{rankblue!80} 1.307
& \cellcolor{rankblue!32} 83.91 & \cellcolor{rankblue!32} 85.01 & \cellcolor{rankblue!32} 79.51 & \cellcolor{rankblue!80} 0.75 & \cellcolor{rankblue!32} 0.53 & \cellcolor{rankblue!14} 1.272 \\
\bottomrule
\end{tabular}
}
% \vspace{0.2em}
\vspace{-1em}
\end{table*}

\FloatBarrier

\subsubsection{Complete Pipeline Results}
Table~\ref{tab:pipeline_full} reports the full per-pipeline metrics for the five
distillation pipelines in Fig.~\ref{fig:pipeline_compare}, both at the
student-initialization stage and after DMD refinement. All pipelines use
Wan2.1-14B as the DMD teacher and its sampled distillation data as the
initialization target; precision and coverage are computed with V-JEPA2 features
against the Wan2.1-14B teacher, and diversity is the raw Vendi score. Consistent
with the analysis in Sec.~\ref{method}, the mode-covering Causal CD initialization
(E) attains the highest initialization and final coverage, whereas mode-seeking
initializations such as Causal DMD (C) reach high precision but lower coverage.

\begin{table}[!htbp]
\centering
\captionsetup{font=footnotesize}
\vspace{-0.21em}
\caption{
\textbf{Complete results across distillation pipelines.}
VBench, precision, coverage, and diversity for the five initialization pipelines
in Fig.~\ref{fig:pipeline_compare}, at the student-initialization stage and after
DMD refinement. For initialization coverage we report both raw coverage
(on the raw outputs) and teacher-normalized re-noised coverage (after the shared
refinement with $\rho=0.9$); all coverage is computed against the Wan2.1-14B
reference. In every column, darker shading indicates a larger value.
}
\vspace{-0.2em}
\label{tab:pipeline_full}
\scriptsize
\setlength{\tabcolsep}{3.5pt}
\renewcommand{\arraystretch}{1.1}
\begin{tabular}{l cccc cccc}
\toprule
\multirow{2}{*}{Pipeline Init.}
& \multicolumn{4}{c}{Student Initialization}
& \multicolumn{4}{c}{Distillation Result} \\
\cmidrule(lr){2-5}\cmidrule(lr){6-9}
& VBench $\uparrow$ & Precision $\uparrow$ & Raw Coverage $\uparrow$ & Re-noised Coverage $\uparrow$
& VBench $\uparrow$ & Precision $\uparrow$ & Coverage $\uparrow$ & Diversity $\uparrow$ \\
\midrule
(A) Bid
& \cellcolor{rankblue!14} 59.09 & \cellcolor{rankblue!14} 0.000 & \cellcolor{rankblue!14} 0.000 & \cellcolor{rankblue!14} 0.000
& \cellcolor{rankblue!14} 80.86 & \cellcolor{rankblue!14} 0.137 & \cellcolor{rankblue!14} 0.055 & \cellcolor{rankblue!14} 1.197 \\
(B) AR
& \cellcolor{rankblue!32} 67.41 & \cellcolor{rankblue!32} 0.238 & \cellcolor{rankblue!32} 0.008 & \cellcolor{rankblue!32} 0.164
& \cellcolor{rankblue!32} 84.48 & \cellcolor{rankblue!56} 0.691 & \cellcolor{rankblue!32} 0.348 & \cellcolor{rankblue!32} 1.247 \\
(C) Causal DMD
& \cellcolor{rankblue!100} 82.41 & \cellcolor{rankblue!80} 0.645 & \cellcolor{rankblue!80} 0.324 & \cellcolor{rankblue!56} 0.324
& \cellcolor{rankblue!56} 84.51 & \cellcolor{rankblue!32} 0.672 & \cellcolor{rankblue!56} 0.422 & \cellcolor{rankblue!56} 1.253 \\
(D) ODE
& \cellcolor{rankblue!56} 71.24 & \cellcolor{rankblue!56} 0.488 & \cellcolor{rankblue!56} 0.020 & \cellcolor{rankblue!80} 0.326
& \cellcolor{rankblue!80} 84.62 & \cellcolor{rankblue!100} 0.867 & \cellcolor{rankblue!80} 0.484 & \cellcolor{rankblue!80} 1.265 \\
(E) Causal CD
& \cellcolor{rankblue!80} 81.85 & \cellcolor{rankblue!100} 0.770 & \cellcolor{rankblue!100} 0.664 & \cellcolor{rankblue!100} 0.648
& \cellcolor{rankblue!100} 84.74 & \cellcolor{rankblue!80} 0.750 & \cellcolor{rankblue!100} 0.527 & \cellcolor{rankblue!100} 1.272 \\
\bottomrule
\end{tabular}
\vspace{-1em}
\end{table}
\vspace{-0.21em}
\subsubsection{Re-noising}
\begin{figure}[!htbp]
\centering
\includegraphics[width=0.70\linewidth]{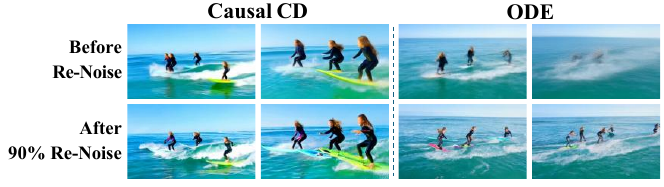}
\captionsetup{font=footnotesize}
\caption{
\textbf{Effect of teacher-normalized re-noising.}
We compare raw initializer samples before and after applying a shared teacher
refinement. Re-noising preserves the initializer-induced scene semantics while
reducing nuisance variation in low-level fidelity, making the distributional
comparison less sensitive to raw sharpness differences.
}
\label{fig:teacher_normalized_renoise}
\end{figure}

\vspace{-0.21em}
\label{sec:renoise_ablation}
In our experiments, evaluating initializers without re-noising lets low-level
fidelity confound coverage: a visibly blurrier initializer such as ODE
(Fig.~\ref{fig:teacher_normalized_renoise}) is scored with very low coverage
despite reaching broad semantic support. In Table~\ref{tab:pipeline_full}, ODE
obtains only $0.020$ raw coverage---far below the sharper Causal DMD
($0.324$)---yet yields \emph{higher} coverage after DMD ($0.484$ vs.\ $0.422$).
Teacher-normalized re-noising removes this misleading gap: ODE reaches $0.326$,
consistent with the post-DMD ordering. This indicates that raw coverage is
dominated by denoising state, whereas the re-noised protocol better reflects the
semantic support available for subsequent DMD refinement.

\FloatBarrier

\section{Related Work}
\label{related work}

\subsection{Autoregressive Video Diffusion Models.}
With the rapid progress of video generation~\citep{wan2025wan,yang2024cogvideox,li2026dynamicsboost,gao2025charactershot,li2024towards}, a growing line of work reformulates video generation autoregressively, either by changing the temporal denoising process itself or by converting a pretrained bidirectional video diffusion model into a causal generator. Diffusion Forcing and progressive autoregressive diffusion introduce per-frame or progressively scheduled noise levels to bridge next-token prediction and full-sequence diffusion \citep{chen2024diffusionforcing,xie2025progressive}. More closely related to our setting, CausVid distills bidirectional video diffusion models into few-step autoregressive generators with ODE initialization and distribution matching \citep{yin2025causvid}; Self-Forcing reduces exposure bias by training on self-generated histories with video-level distribution matching \citep{huang2025selfforcing}; Causal Forcing shows that ODE initialization from a bidirectional teacher can be target-misaligned and instead uses an autoregressive teacher before DMD refinement \citep{zhu2026causal}; and Causal Forcing++ further scales this idea with causal consistency distillation for frame-wise one- and two-step generation \citep{zhao2026causalforcingpp}. Recent follow-ups push this direction toward longer horizons, stronger controls, and lower latency: Rolling Forcing, Self-Forcing++, Context Forcing, and MotionStream target multi-minute rollout or interactive motion control \citep{liu2025rolling,cui2026selfforcingpp,chen2026contextforcing,shin2026motionstream}; and HiAR and BiWM explore hierarchical or bidirectional autoregressive generation to mitigate accumulated errors \citep{zou2026hiar,rui2026biwm}. %These works mainly optimize latency, stability, and perceptual quality. In contrast, our work focuses on a complementary question: whether the autoregressive student initialization actually covers the conditional teacher distribution before reverse-KL-style refinement.

\subsection{Forward and Reverse Divergence in Diffusion Models.}
A useful lens for diffusion distillation is the divergence implicitly induced by the training distribution. Offline, teacher-supervised objectives---including progressive distillation, consistency distillation, consistency trajectory models, and continuous-time consistency models---train on teacher or data trajectories and thus exhibit forward-divergence or mean-seeking behavior, which tends to cover teacher samples but may average modes or lose fine detail \citep{salimans2022progressive,song2023consistency,kim2024ctm,lu2025scm}. In contrast, on-policy score-distillation and distribution-matching objectives, including DMD, DMD2, and SiD, optimize student-generated samples against teacher scores and are commonly associated with reverse-KL-style, mode-seeking behavior: they improve sharpness and fidelity but can collapse onto high-probability modes \citep{yin2024one,yin2024improved,zhou2024sid}. Recent work has made this trade-off explicit. (f)-Distill generalizes score-based distribution matching beyond reverse KL and shows that forward KL or Jensen--Shannon alternatives can improve mode coverage \citep{xu2025fdistill}; rCM combines the mode-covering behavior of continuous-time consistency with score regularization to recover reverse-divergence fidelity at large scale \citep{zheng2026rcm}; ADM/DMDX uses adversarial distribution matching to alleviate the mode collapse caused by reverse-KL DMD \citep{lu2025adm}; SGMD analyzes the mode-seeking conservatism of DMD-style video distillation and improves motion dynamics by matching score gradients \citep{wu2026sgmd}; Adaptive Video Distillation targets oversaturation, temporal inconsistency, and temporal collapse in few-step video generation \citep{you2026adaptive}; and Mode Seeking meets Mean Seeking decouples a supervised mean-seeking flow-matching head from a reverse-KL distribution-matching head for long-video synthesis \citep{cai2026mode}. In autoregressive video generation, HiAR explicitly adds a forward-KL regularizer to preserve motion diversity under self-rollout, and BiWM adds GAN and mass-covering forward-KL objectives to counter DMD-induced mode-seeking degradation \citep{zou2026hiar,rui2026biwm}. %Our method follows this forward--reverse complementarity, but emphasizes teacher-aligned coverage: the forward-KL stage should match the same conditional teacher distribution used by the subsequent DMD stage, rather than serving merely as a generic causal warm start.

\section{Conclusion}
\label{Conclusion}
% \section{Conclusion}

We revisit autoregressive video distillation from a distributional perspective.
We show that existing multi-stage pipelines often decouple pre-DMD initialization from DMD refinement and evaluate initialization mainly by visual scores.
However, trajectory- or consistency-based initialization primarily provides mode coverage, while DMD performs mode-seeking (reverse-KL) refinement, suggesting that effective initialization should provide matched mode coverage for the target DMD teacher.
Based on this observation, we introduce a distributional evaluation protocol and further show that pure DMD can still induce late-stage distribution drift even under aligned targets.
We address this with joint distillation, combining DMD's mode-seeking objective with CD's mode-covering constraint.
Experiments demonstrate that our method achieves a better balance among visual quality, coverage, and diversity, highlighting the importance of distributional alignment in autoregressive video distillation.

\bibliography{iclr2025_conference}
\bibliographystyle{iclr2025_conference}

\appendix
\section{More Details of Coverage Metrics}
\label{app:coverage_metrics}
This section specifies the complete evaluation pipeline used for the precision,
coverage, and diversity results in the paper. We distinguish the
\emph{normalization teacher} $T_{\mathrm{norm}}$, which removes nuisance
differences among raw initializer outputs, from the \emph{reference teacher}
$T_{\mathrm{ref}}$, whose sample distribution defines the target support. The
two teachers need not be identical. This distinction is important: the
normalization operation is shared across initializers, whereas
$T_{\mathrm{ref}}$ is selected according to the target DMD teacher reported in
each experiment.

\paragraph{Evaluation sample sets.}
We use a fixed grid of 16 prompts and 16 random seeds, yielding one video for
each prompt--seed pair and $N=M=256$ videos for both the evaluated model and the
reference teacher. The seeds are
\[
\{11,22,33,42,44,55,66,77,88,123,456,789,2024,3407,7777,9999\}.
\]
The same grid is used for every method and teacher. Matching the grid controls
the prompt mixture and sampling budget, but the reported precision and coverage
remain set-level statistics: they are computed on the pooled 256-sample sets,
not by comparing only videos that share a prompt or seed. We require exactly
one sample for every key $(\text{prompt},\text{seed})$ and do not impute failed
or missing generations.

\paragraph{Teacher-normalized refinement of initializer samples.}
For a pre-DMD initializer, we reuse its saved terminal video latent
$z_{\theta,0}$; this avoids an additional decode--encode pass. Following the
linear path used by the flow-matching teacher~\citep{lipman2022flow}, we draw
$\epsilon\sim\mathcal{N}(0,I)$ and form
\begin{equation}
    z_{\theta,\rho}
    = (1-\rho)z_{\theta,0}+\rho\epsilon.
    \label{eq:appendix_renoise}
\end{equation}
For a given prompt--seed pair, the noise seed is fixed and shared across
initializers, so the normalization itself does not introduce method-dependent
randomness. Unless otherwise stated, $\rho=0.9$. We then apply a fixed
Wan2.1-T2V-14B normalization teacher with classifier-free guidance scale 5.0.
All methods use the same unconditional negative prompt. The solver is
Flow-UniPC~\citep{zhao2023unipc}, configured with 1,000 training timesteps,
schedule shift 1, and no dynamic shifting. We use
$22=\operatorname{round}(25\rho)$ model evaluations. Their sigma values are
uniformly spaced from $0.9$ through $0.9/22$, after which the scheduler performs
the terminal update to zero. The refined latent is decoded at 16 fps to obtain
$\tilde{x}_{\theta}^{(0.9)}$.

The reference set is generated separately with the 25-step sampler of
$T_{\mathrm{ref}}$. Thus, for experiments targeting Wan2.1-14B, the shared
1.3B model performs only the normalization step, while the neighborhood radii
and all reported precision/coverage values are defined by Wan2.1-14B reference
samples. Post-DMD generators are evaluated directly from their
final videos without re-noising: their outputs are already fully denoised, and
the purpose of normalization is specifically to remove the heterogeneous
denoising state of pre-DMD initializers.

\paragraph{Temporal standardization and video representation.}
Our default distribution protocol uses the frozen
\texttt{facebook/vjepa2-vith-fpc64-256} V-JEPA2 encoder
\citep{assran2025vjepa2}. We restrict every video to the first five seconds. In
our 16-fps setup this is the 81-frame prefix; for methods that generate longer
videos, later frames are discarded before feature extraction. We select eight
frames with rounded, uniformly spaced indices over this prefix, including both
temporal endpoints. The same frame-selection rule is applied to teacher and
student videos. This
short-window protocol is applied symmetrically to the student and reference
sets, and its absolute values are not mixed with the default 8-frame results in
the main comparison.

The selected frames are processed by the model's native video processor and
encoded in bfloat16. Let $h_1,\ldots,h_L\in\mathbb{R}^{1280}$ denote the output
spatiotemporal tokens. We aggregate them into one video descriptor by
concatenating the token-wise mean and sample standard deviation (correction
one, matching the extractor implementation),
\begin{equation}
    f(x)=\left[
        \frac{1}{L}\sum_{\ell=1}^{L}h_\ell\ ;\
        \operatorname{std}_{\ell=1}^{L}(h_\ell)
    \right]\in\mathbb{R}^{2560},
    \qquad
    \phi(x)=\frac{f(x)}{\lVert f(x)\rVert_2}.
    \label{eq:appendix_vjepa_feature}
\end{equation}
All reported results use these raw, $\ell_2$-normalized V-JEPA2 features. We do
not apply low-pass filtering, sharpness residualization, dimensionality
reduction, or PCA before computing the metrics. PCA is used only for the
visualizations in Fig.~\ref{fig:coverage_evolution} and does not affect any
reported number.

\paragraph{Local teacher support.}
We use a non-parametric $k$-nearest-neighbor support estimator, following the
improved precision construction of \citet{kynkaanniemi2019improved} and the
coverage statistic of \citet{naeem2020reliable}. Let
$V=\{v_j\}_{j=1}^{M}$ be the normalized features of $T_{\mathrm{ref}}$ and
$U=\{u_i\}_{i=1}^{N}$ those of the evaluated model. Distances are Euclidean in
the normalized feature space,
\begin{equation}
    d(a,b)=\lVert a-b\rVert_2,
    \qquad
    d(a,b)^2=2-2a^\top b,
\end{equation}
so ranking by this distance is equivalent to ranking by cosine distance. For
each teacher feature, we define an adaptive local radius
\begin{equation}
    r_k(v_j)
    = \text{distance from }v_j\text{ to its }k\text{-th nearest element of }
      V\setminus\{v_j\}.
    \label{eq:appendix_teacher_radius}
\end{equation}
We use $k=5$ for every experiment; $k$ is fixed globally and is never tuned per
method, checkpoint, or teacher.

\paragraph{Precision and coverage.}
Precision is the fraction of evaluated samples that fall inside at least one
adaptive teacher neighborhood:
\begin{equation}
    \operatorname{Prec}(U,V)
    =\frac{1}{N}\sum_{i=1}^{N}
      \mathbf{1}\!\left[
      \exists j:\ d(u_i,v_j)\leq r_k(v_j)
      \right].
    \label{eq:appendix_precision}
\end{equation}
Coverage is the fraction of teacher samples whose own neighborhood contains at
least one evaluated sample:
\begin{equation}
    \operatorname{Cov}(U,V)
    =\frac{1}{M}\sum_{j=1}^{M}
      \mathbf{1}\!\left[
      \min_i d(v_j,u_i)\leq r_k(v_j)
      \right].
    \label{eq:appendix_coverage}
\end{equation}
Although reported beside precision, this coverage statistic is not the recall
of the improved precision--recall metric. Recall constructs neighborhoods from
the generated samples; Eq.~\ref{eq:appendix_coverage} deliberately uses only
teacher radii, which makes the target support fixed across all compared methods
and prevents isolated student outliers from creating artificially large
regions. With $M=256$, coverage changes in increments of $1/256$ before
rounding.

The two statistics measure complementary failure modes. High precision means
that most model samples lie near the estimated teacher support, but it does not
imply broad support: a strongly mode-seeking generator can have high precision
and low coverage. High coverage means that many local teacher neighborhoods are
reached, but it is not a standalone perceptual-quality score. Both quantities
are conditional on the feature encoder, prompt set, temporal window, and
$T_{\mathrm{ref}}$. Consequently, values computed against different reference
teachers should be interpreted relative to their respective supports rather
than ranked as if they shared one absolute scale.

\paragraph{Diversity statistic.}
For completeness, all diversity entries in the paper use the raw Vendi score
\citep{friedman2023vendi} on the same normalized V-JEPA2 descriptors. For a
method feature matrix $U\in\mathbb{R}^{N\times2560}$, we form the shifted cosine
kernel
\begin{equation}
    K=\frac{UU^\top+\mathbf{1}\mathbf{1}^\top}{2},
    \qquad \bar K=\frac{K}{N}.
\end{equation}
If $\{\lambda_q\}$ are the positive eigenvalues of $\bar K$, the reported score
is
\begin{equation}
    \operatorname{Vendi}(U)
    =\exp\!\left(-\sum_q\lambda_q\log\lambda_q\right).
\end{equation}
We report this raw effective-rank value, not a ratio to the teacher. It is a
within-method diversity statistic and does not enter the precision or coverage
calculation.

\section{Additional Ablation}
\label{app:complete_ablation}
\subsection{Joint Distillation}
To complement the Wan2.1-14B ablations in Sec.~\ref{sec:joint_ablation}, we
repeat the joint distillation ablations on the weak setting, where Wan2.1-1.3B
serves as both the initialization data source and the DMD teacher. The
conclusions are consistent with the main setting: joint distillation better
preserves coverage and diversity than pure DMD, and $\lambda=0.01$ provides the
best overall trade-off.

\paragraph{Loss weight $\lambda$.}
Table~\ref{tab:lambda_ablation_1p3b} ablates $\lambda$ under the weak setting,
reporting the best checkpoint within a 2500-step training trajectory for each
$\lambda$ as in the main setting. As $\lambda$ decreases, the DMD mode-seeking
effect strengthens, improving precision but reducing coverage and diversity;
$\lambda=0$ recovers pure DMD. Consistent with the Wan2.1-14B result in
Tab.~\ref{tab:lambda_ablation}, $\lambda=0.01$ gives the best overall trade-off
among quality, precision, and coverage.

\begin{table}[!htbp]
\centering
\captionsetup{font=footnotesize}
\caption{
\textbf{Ablation on the joint loss weight $\lambda$ under the weak (Wan2.1-1.3B) setting.}
A larger $\lambda$ strengthens the CD-based mode-covering constraint, while a
smaller $\lambda$ gives more weight to DMD refinement, and $\lambda=0$ recovers
pure DMD. Diversity is reported as the raw Vendi score under the same V-JEPA2
protocol as Tab.~\ref{tab:main_comparison}. In each column, darker shading indicates a larger value.
}
\label{tab:lambda_ablation_1p3b}
\footnotesize
\setlength{\tabcolsep}{5.0pt}
\renewcommand{\arraystretch}{1.05}
\begin{tabular}{c ccc ccc}
\toprule
\multirow{2}{*}{$\lambda$}
& \multicolumn{3}{c}{VBench Metrics $\uparrow$}
& \multicolumn{3}{c}{Distributional Metrics $\uparrow$} \\
\cmidrule(lr){2-4}\cmidrule(lr){5-7}
& Total & Quality & Semantic & Precision & Coverage & Diversity \\
\midrule
1
& \cellcolor{rankblue!14} 84.18 & \cellcolor{rankblue!14} 85.17 & \cellcolor{rankblue!56} 80.18 & \cellcolor{rankblue!14} 0.65 & \cellcolor{rankblue!80} 0.53 & \cellcolor{rankblue!32} 1.295 \\
0.1
& \cellcolor{rankblue!32} 84.34 & \cellcolor{rankblue!32} 85.41 & \cellcolor{rankblue!14} 80.06 & \cellcolor{rankblue!32} 0.67 & \cellcolor{rankblue!56} 0.52 & \cellcolor{rankblue!56} 1.298 \\
0.01
& \cellcolor{rankblue!100} 84.54 & \cellcolor{rankblue!80} 85.61 & \cellcolor{rankblue!100} 80.28 & \cellcolor{rankblue!100} 0.77 & \cellcolor{rankblue!100} 0.59 & \cellcolor{rankblue!100} 1.305 \\
0.001
& \cellcolor{rankblue!80} 84.52 & \cellcolor{rankblue!100} 85.63 & \cellcolor{rankblue!32} 80.09 & \cellcolor{rankblue!56} 0.71 & \cellcolor{rankblue!32} 0.50 & \cellcolor{rankblue!80} 1.299 \\
0
& \cellcolor{rankblue!56} 84.50 & \cellcolor{rankblue!56} 85.58 & \cellcolor{rankblue!80} 80.19 & \cellcolor{rankblue!80} 0.72 & \cellcolor{rankblue!14} 0.44 & \cellcolor{rankblue!14} 1.275 \\
\bottomrule
\end{tabular}

\vspace{0.2em}
\end{table}

\paragraph{Across training steps.}
Table~\ref{tab:joint_ablation_1p3b} tracks joint distillation and pure DMD across
training steps under the weak setting. Pure DMD steadily loses coverage and
diversity as training proceeds, whereas joint distillation maintains higher
coverage and diversity while reaching comparable or better total scores in the
later stage, mirroring the Wan2.1-14B trend in Tab.~\ref{tab:joint_ablation}.

\begin{table}[!htbp]
\centering
\captionsetup{font=footnotesize}
\caption{
\textbf{Ablation of joint distillation across training steps under the weak
(Wan2.1-1.3B) setting.}
Compared with pure DMD, joint distillation better preserves coverage and
diversity while achieving comparable or better VBench scores. Diversity is
reported as the raw Vendi score under the same V-JEPA2 protocol as
Tab.~\ref{tab:main_comparison}. In each column, darker shading indicates a larger value.
}
\label{tab:joint_ablation_1p3b}
\scriptsize
\setlength{\tabcolsep}{3.0pt}
\renewcommand{\arraystretch}{1.08}
\resizebox{\textwidth}{!}{
\begin{tabular}{c ccc ccc @{\hspace{0.8em}} ccc ccc}
\toprule
\multirow{3}{*}{Step}
& \multicolumn{6}{c}{Joint Distillation ($\lambda=0.01$)}
& \multicolumn{6}{c}{Pure DMD ($\lambda=0$)} \\
\cmidrule(lr){2-7}\cmidrule(lr){8-13}
& \multicolumn{3}{c}{VBench Metrics $\uparrow$} & \multicolumn{3}{c}{Distributional Metrics $\uparrow$}
& \multicolumn{3}{c}{VBench Metrics $\uparrow$} & \multicolumn{3}{c}{Distributional Metrics $\uparrow$} \\
\cmidrule(lr){2-4}\cmidrule(lr){5-7}\cmidrule(lr){8-10}\cmidrule(lr){11-13}
& Total & Quality & Semantic & Precision & Coverage & Diversity
& Total & Quality & Semantic & Precision & Coverage & Diversity \\
\midrule
0
& \cellcolor{rankblue!14} 82.16 & \cellcolor{rankblue!14} 83.38 & \cellcolor{rankblue!14} 77.29 & \cellcolor{rankblue!14} 0.61 & \cellcolor{rankblue!56} 0.52 & \cellcolor{rankblue!80} 1.301
& \cellcolor{rankblue!14} 82.16 & \cellcolor{rankblue!32} 83.38 & \cellcolor{rankblue!14} 77.29 & \cellcolor{rankblue!32} 0.61 & \cellcolor{rankblue!100} 0.52 & \cellcolor{rankblue!100} 1.301 \\

500
& \cellcolor{rankblue!32} 83.84 & \cellcolor{rankblue!32} 85.07 & \cellcolor{rankblue!32} 78.92 & \cellcolor{rankblue!32} 0.66 & \cellcolor{rankblue!56} 0.52 & \cellcolor{rankblue!32} 1.295
& \cellcolor{rankblue!56} 84.44 & \cellcolor{rankblue!80} 85.52 & \cellcolor{rankblue!56} 80.09 & \cellcolor{rankblue!56} 0.68 & \cellcolor{rankblue!80} 0.48 & \cellcolor{rankblue!80} 1.293 \\

1000
& \cellcolor{rankblue!56} 84.17 & \cellcolor{rankblue!80} 85.38 & \cellcolor{rankblue!56} 79.29 & \cellcolor{rankblue!56} 0.69 & \cellcolor{rankblue!32} 0.51 & \cellcolor{rankblue!14} 1.292
& \cellcolor{rankblue!80} 84.48 & \cellcolor{rankblue!80} 85.52 & \cellcolor{rankblue!100} 80.35 & \cellcolor{rankblue!80} 0.72 & \cellcolor{rankblue!56} 0.46 & \cellcolor{rankblue!56} 1.284 \\

1500
& \cellcolor{rankblue!80} 84.38 & \cellcolor{rankblue!56} 85.33 & \cellcolor{rankblue!100} 80.58 & \cellcolor{rankblue!80} 0.72 & \cellcolor{rankblue!80} 0.53 & \cellcolor{rankblue!56} 1.297
& \cellcolor{rankblue!100} 84.50 & \cellcolor{rankblue!100} 85.58 & \cellcolor{rankblue!80} 80.19 & \cellcolor{rankblue!80} 0.72 & \cellcolor{rankblue!32} 0.44 & \cellcolor{rankblue!32} 1.275 \\

2500
& \cellcolor{rankblue!100} 84.54 & \cellcolor{rankblue!100} 85.61 & \cellcolor{rankblue!80} 80.28 & \cellcolor{rankblue!100} 0.77 & \cellcolor{rankblue!100} 0.59 & \cellcolor{rankblue!100} 1.305
& \cellcolor{rankblue!32} 84.09 & \cellcolor{rankblue!56} 85.22 & \cellcolor{rankblue!32} 79.59 & \cellcolor{rankblue!100} 0.73 & \cellcolor{rankblue!14} 0.43 & \cellcolor{rankblue!14} 1.269 \\
\bottomrule
\end{tabular}
}
\end{table}

\section{Prompts}
%  正文每个fig对应的prompt
For reproducibility, we list the prompts used for the qualitative examples and visual comparisons in this paper. For fair comparison, all methods in the same figure use the same prompt, and when applicable, the same set of fixed random seeds. The reference column indicates the figures in which each prompt is used.

\footnotesize
\begin{longtable}{>{\raggedright\arraybackslash}p{0.80\textwidth}
                  >{\centering\arraybackslash}p{0.15\textwidth}}
\caption{Prompts used in this paper.}
\label{tab:used_prompts} \\
\toprule
\textbf{Prompt} & \textbf{Reference} \\
\midrule
\endfirsthead

\toprule
\textbf{Prompt} & \textbf{Reference} \\
\midrule
\endhead

\bottomrule
\endfoot

Four teenage girls, with windswept hair and determined expressions, ride powerful waves in frigid, crystal-clear coldwater. Their vibrant surfboards cut through frothy whitecaps as they lean into turns with athletic grace. Dynamic camera angles show drone sweeps, close-up shoulder shots, and overhead tracking capture their synchronized movements. Each girl wears snug wetsuits, laughing and shouting as they paddle, carve, and ride the swell. Motion is fluid and energetic, emphasizing their youthful spirit against the vast, chilly ocean backdrop.
& Fig.~\ref{fig:teacher_normalized_renoise} \\
\midrule

Zara, a young woman with flowing hair and curious expression, stumbles joyfully upon a small, crystal-clear brook, its surface shimmering with sunlight as it bubbles gently over smooth pebbles visible beneath. She bends slightly, reaching out to touch the cool water, her reflection dancing in the ripples. The camera glides softly around her, capturing her delighted surprise from low angles and overhead shots, emphasizing the serene, natural beauty and the soothing motion of water swirling around her feet.
& Fig.~\ref{fig:pipeline_diversity} \\
\midrule

A young couple stands side by side under a vast, starry night sky, gazing upward in wonder as distant stars flicker with soft, ethereal light. Their faces glow with awe, hands lightly touching, shoulders relaxed. The camera slowly pans around them, tilting upward to capture the infinite cosmos, then gently zooms in on their eyes reflecting the stars. Shot in cinematic 4K HD, with deep blues and warm golden hues, the scene pulses with quiet romance and cosmic wonder as gentle wind rustles their hair.
& Fig.~\ref{fig:pipeline_diversity} \\
\midrule

A vibrant red ball soars dynamically through the air, arcing gracefully toward a graceful lady in a flowing green dress, standing poised in a sunlit field. Her posture is serene, eyes tracking the ball’s approach as it nears her midsection. The camera tracks smoothly from a low angle, rising slightly as the ball descends. Soft wind ripples her dress. The scene glows with cinematic realism, rich colors, and natural motion — the ball’s flight and her subtle lean create anticipation.
& Fig.~\ref{fig:exp_comparison_diversity} \\
\midrule

Sammy, a curious young boy with tousled brown hair and a determined expression, sprints through the dappled forest toward a towering ancient oak. He leaps with joy, sliding headfirst into its hollow trunk as leaves rustle around him. The camera follows him in a dynamic tracking shot, then swoops low to capture his playful disappearance. Sunlight filters through the canopy, casting dancing shadows. His laughter echoes as he vanishes, leaving the forest alive with mystery and motion.
& Fig.~\ref{fig:exp_comparison_diversity} \\
\midrule

A man with long, flowing curly black hair stands alone on a windswept beach cliff, gazing at the sun setting over the endless ocean captured from behind in a stunning anime-style digital illustration by Makoto Shinkai. His hollow cheeks and contemplative posture evoke melancholy beauty, as if frozen in his last moment of solitude. The scene glows with warm, cinematic hues, enhanced by soft atmospheric lighting and gentle waves crashing below. The camera slowly pans right, then tilts up to frame the sky, blending romanticism with aestheticism. Motion: slow, reflective, emotionally resonant.
& Fig.~\ref{fig:exp_comparison} \\
\midrule

The story ends with little boy Teddy, wearing a cozy red sweater and holding a glowing lantern, joyfully dancing beside fairy Sparkle, whose wings shimmer with aurora-like colors. They float side-by-side through a dreamy forest at dusk, laughing and twirling as camera circles them in slow motion. 3D animation, 4K resolution, 16:9, cinematic lighting, soft depth of field, whimsical fantasy style. Their bond glows with warmth as they leap into the next adventure, hand in hand.
& Fig.~\ref{fig:exp_comparison} \\
\midrule

Cinematic wide shot of a tiny tarsier perched dynamically on King Kong’s giant palm, surrounded by lush forest near cascading waterfalls, with fluttering birds soaring in golden sunlight that highlights the misty background. King Kong leaps joyfully with playful companions, capturing natural motion through fluid camera pans and dolly movements. The tarsier’s curious expression and agile movements contrast with the colossal scale of its surroundings, emphasizing whimsy and wonder in a vibrant, high-detail fantasy scene.
& Fig.~\ref{fig:exp_comparison} \\
\midrule

A young, radiant beauty gracefully walks through an opulent luxury room, mid-shot, her flowing gown swaying with each step. Soft golden light bathes her as she glides past gilded furniture and crystal chandeliers, her expression serene yet confident. The camera smoothly follows her from a medium distance, subtly panning to capture her poised posture and elegant stride. Rich textures, ambient reflections, and slow-motion movement enhance the luxurious atmosphere, making every gesture feel graceful and intentional.
& Fig.~\ref{fig:exp_comparison} \\

\end{longtable}

\end{document}